\documentclass[10pt,twocolumn,letterpaper]{article}

\usepackage[review]{cvpr}      % To produce the REVIEW version
\definecolor{cvprblue}{rgb}{0.21,0.49,0.74}
\usepackage[pagebackref,breaklinks,colorlinks,allcolors=cvprblue]{hyperref}
\usepackage{float}
\usepackage{tikz}
\newcommand{\tikzxmark}{%
\tikz[scale=0.15] {
    \draw[line width=0.7,line cap=round] (0,0) to [bend left=6] (1,1);
    \draw[line width=0.7,line cap=round] (0.2,0.95) to [bend right=3] (0.8,0.05);
}}
\newcommand{\tikzcmark}{%
\tikz[scale=0.15] {
    \draw[line width=0.7,line cap=round] (0.25,0) to [bend left=10] (1,1);
    \draw[line width=0.8,line cap=round] (0,0.35) to [bend right=1] (0.23,0);
}}
\def\paperID{9240} % *** Enter the Paper ID here
\def\confName{CVPR}

\title{MambaDETR: Query-based Temporal Modeling using State Space Model for Multi-View 3D Object Detection}

\author{Tong Ning\\
Institution1\\
Institution1 address\\
{\tt\small firstauthor@i1.org}
\and
Second Author\\
Institution2\\
First line of institution2 address\\
{\tt\small secondauthor@i2.org}
}

\begin{document}
\maketitle
\begin{abstract}
Utilizing temporal information to improve the performance of 3D detection has made great progress recently in the field of autonomous driving. Traditional transformer-based temporal fusion methods suffer from quadratic computational cost and information decay as the length of the frame sequence increases. In this paper, we propose a novel method called MambaDETR, whose main idea is to implement temporal fusion in the efficient state space. Moreover, we design a Motion Elimination module to remove the relatively static objects for temporal fusion. On the standard nuScenes benchmark, our proposed MambaDETR achieves remarkable result in the 3D object detection task, exhibiting state-of-the-art performance among existing temporal fusion methods.
\end{abstract}    
\section{Introduction}
\label{sec:intro}

Multi-view 3D object detection is a fundamental task in autonomous driving, enabling vehicles to perceive their surrounding environment using sensor data. Recent studies have leveraged temporal information from image frame sequences to improve detection performance. Transformers \cite{vaswani2017attention} with attention mechanisms have proven effective in modeling dependencies for sequential inputs, leading many approaches to adopt transformer-based temporal fusion methods to explore temporal information for 3D detection.

\begin{figure}
\centering
\includegraphics[width=.98\linewidth]{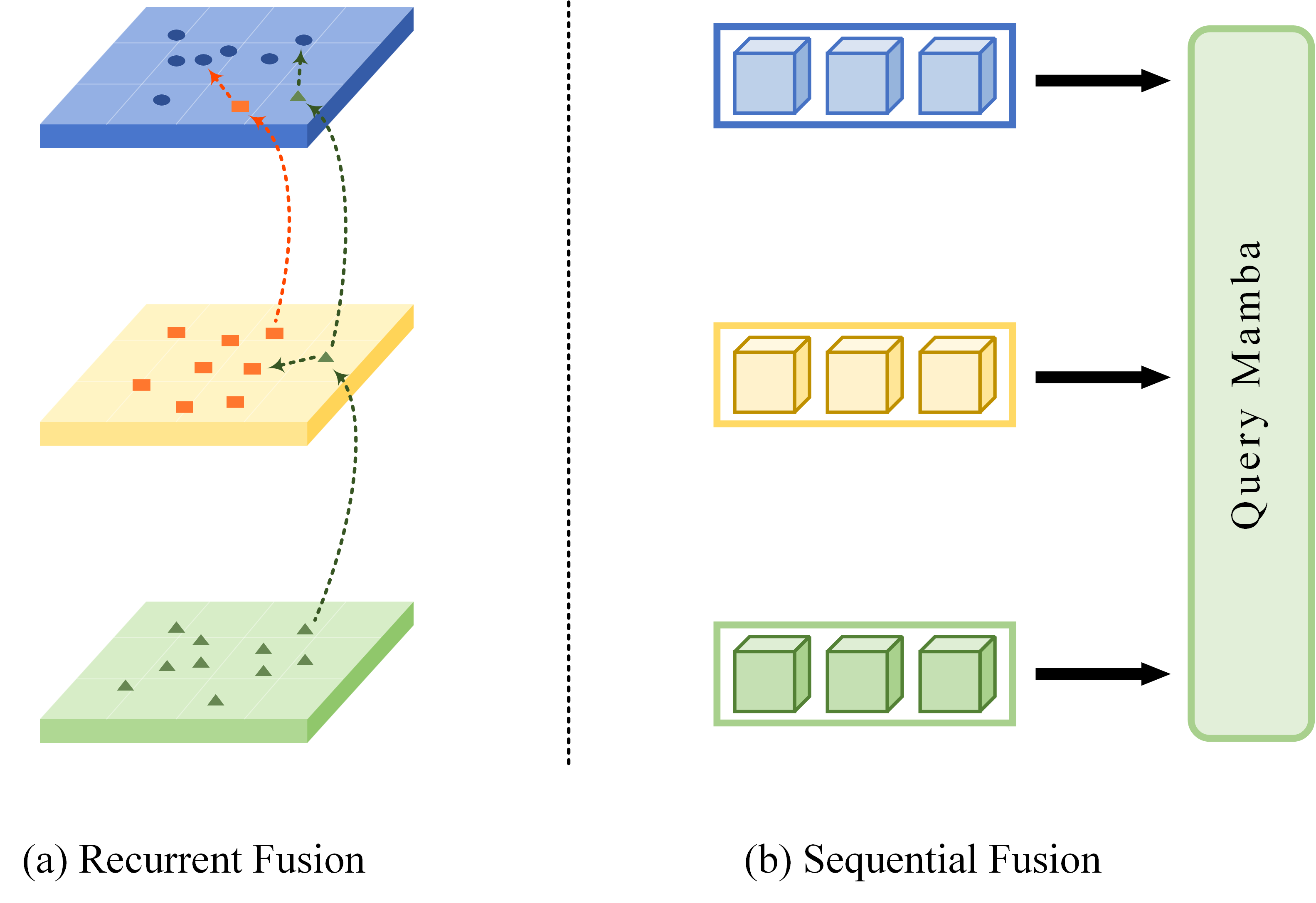} % Reduce the figure size so that it is slightly narrower than the column. Don't use precise values for figure width.This setup will avoid overfull boxes.
% \captionsetup{justification=centerlast} % Center the caption
\caption{
Different temporal fusion methods in recurrent manner and sequential manner.
% Overall architecture of DinoQuery. The image backbone network extracts features from the input multi-view images, which are then shared with the Heatmap Head and Grounding DINO modules. The Heatmap Head generates 2D global queries, while Grounding DINO produces 2D category-customized queries based on visual and textual prompts. These queries are combined to form a comprehensive set of 2D queries, which are then processed by the 3D Query Generator. This generator constructs 3D queries that are refined and passed through the decoder layers to predict the 3D detection boxes.
}
\label{cover}
\end{figure}

Existing transformer-based temporal fusion methods, such as \cite{liu2023petrv2, luo2022detr4d}, use one adjacent historical frame to interact with the current frame in the transformer decoder, which can efficiently improve 3D detection performance. However, these methods suffer from a quadratic increase in computational cost as the sequence length grows, limiting their ability to adopt more frames for temporal interaction.
To solve this problem, subsequent methods \cite{lin2022sparse4d, wang2023exploring, li2022bevformer, cai2022memot, he2022queryprop, zeng2022motr, meinhardt2022trackformer, jiang2019video} incorporate multiple frames into the temporal fusion module by fusing frames in a recurrent (See Figure \ref{cover} a) rather than a sequential manner. The long-term historical information is propagated frame by frame, allowing each frame’s features to integrate information from preceding frames. However, this recurrent fusion process can cause information decay over time, leading the model to focus more on current information rather than long-term frame \cite{lin2022sparse4d, li2022bevformer}.

% The sequential fusion method \cite{Bevdet4d, park2022time, wang2023focal} accumulates features from historical frames to enhance the current frame's feature representation through concatenation. Such concatenation-based temporal fusion are simple and extendable

% however, the fused features retain information only corresponding to their own timestamps. As the number of frames increases, the computational costs grow linearly, limiting the effective exploitation of long-range information.

% In contrast, the recurrent fusion method \cite{hou2024query, wang2023exploring} propagates long-term historical information frame by frame, enabling each frame's features to incorporate historical data from preceding frames. However, this approach experiences information decay over time during the recurrent propagation, leading the model to focus more on current information and less on long-term data. 

To address these challenges, we propose a novel method called MambaDETR, which represents temporal fusion candidates as 3D queries initialized from 2D proposals and performs temporal fusion in a hidden space using an SSM-based (Structured State Space Model) module, as shown in Figure \ref{cover} (b). This approach replaces the traditional transformer-based module with an SSM-based module, enabling temporal fusion in a sequential manner that effectively models long-range information while maintaining only linear memory and computation costs.
Specifically, given a sequence of frames, we first employ a 2D detector to independently generate high-quality 2D proposals for each frame. These proposals are then used to produce 3D object queries through 3D projection, as in previous methods \cite{jiang2024far3d}, resulting in a sequence of 3D object queries at each time step. Furthermore, according to the physical laws of motion, the same 3D object does not significantly shift between adjacent frames. Therefore, fusing all the queries from the adjacent frames can lead to unnecessary computational costs. 
Based on this insight, we introduce a Motion Elimination Module, which aligns the objects from the previous frame to the current frame through ego transformation and generates a motion mask based on the relative movements of objects. Consequently, trivial and relative static objects in the ego vehicle coordinate are removed from the previous frames, retaining only the moving objects and thereby enabling more efficient temporal fusion.

% Based on this insight, we introduce a Motion Elimination Module, which removes redundant, motionless queries, retaining only moving queries based on the explicit position. 

% The motion elimination module align the objects from the previous frame to the current frame through ego transformation and generate mask by the explicit position of objects.

The refined sequence of 3D object queries is then input into the Query Mamba Module, which performs query-based temporal fusion in the state space. By leveraging a structured state space layer, the Query Mamba Module achieves long-range modeling without pairwise comparisons. Consequently, MambaDETR can be efficiently applied to long-range sequences of image frames.

In summary, our contributions include:

\begin{itemize}
\item A novel temporal fusion method called MambaDETR is proposed for 3D object detection, which achieves efficient temporal fusion in the state space. This approach sequentially fuses frame sequences, thereby fully exploiting long-range information while avoid quadratic complexity.

\item The Motion Elimination module is designed, which removes relatively static objects while retaining moving ones in the ego vehicle coordinates, thereby improving fusion efficiency and reducing computational cost.

\item Comprehensive experiments have been conducted on the standard nuScenes dataset, and the evaluation results demonstrate the superior performance of MambaDETR in 3D object detection, with linear computation cost compared to transformer-based methods, e.g. StreamPETR. 
\end{itemize}

% To tackle this problem, researchers propose a novel State Space Model(SSM), which demonstrate effective performance in long sequence modeling with linear complexity.

% Inspired by this advancement of SSM-based architectures, we introduce QMamba, a query based 3D object detection approach that integrating SSM-based module to facilitate efficient temporal fusion. Given the input of frame sequence, we first use 2D detector to obtain 2D proposals from each frame independently. For each frame, we leverage the high-quality 2D proposals to generate 3D object queries through 3D projection. 

% We concatenate the all the 3D object queries within the same frame together with the motion embedding and build the \textcolor{blue}{temporal fusion candidates query}. The \textcolor{blue}{temporal fusion candidates query} sequence are treated as the input to the \textcolor{blue}{Mamba module}, which can realize query based temporal fusion in the hidden states. 

\section{Related Work}
\label{sec:related_work}

\subsection{Temporal Modeling in Multi-view 3D Object Detection}

Multi-view 3D detection is a crucial task in autonomous driving, requiring the processing of multiple camera images and the prediction of 3D bounding boxes. 
Pioneer's research~\cite{wang2022detr3d,petr_liu_2022,huang2021bevdet,li2022bevformer,jiang2023polarformer,wang2023object} centers on the effective conversion of various perspective views into a cohesive 3D space within a single frame. The transformation process can be categorized into two groups: methods based on dense BEV (Bird's Eye View) representation ~\cite{huang2021bevdet,li2022bevformer,huang2023fast,li2023bevdepth,jiang2023polarformer, liu2023bevfusion, liang2022bevfusion} and methods based on sparse queries~\cite{wang2022detr3d,petr_liu_2022,lin2022sparse4d,wang2023object}. 

Recent studies~\cite{li2022bevformer,sts_wang_2022,huang2022bevdet4d,park2022time,lin2022sparse4d,luo2022detr4d,cape_xiong_2023,liBevstereoEnhancingDepth2023,wang2023exploring} have incorporated temporal information to address the occlusion issue and enhance speed prediction accuracy.  
BEVFormer \cite{li2022bevformer} first introduces sequential temporal modeling into multi-view 3D object detection and applies temporal self-attention. The BEVDet4D \cite{huang2022bevdet4d} paradigm is proposed to be lifted from the spatial-only 3D space to the temporal 4D space. For the first time, vision-based methods become comparable with those using radar or LiDAR.  
%STS \cite{sts_wang_2022} proposed the Surround-view Temporal Stereo technique, leveraging geometry correspondence between frames across time to enhance depth learning for improved 3D detection. PETR \cite{petr_liu_2022} proposed the Position Embedding Transformation technique for multi-view 3D object detection, which allows object queries to extract position-aware features in 3D space. 
Sparse4D \cite{lin2022sparse4d} refines anchor boxes iteratively by incorporating sparse spatial-temporal fusion to improve sparse 3D detection. Additionally, DETR4D \cite{luo2022detr4d} introduces a novel hybrid approach that performs cross-frame fusion over past object queries and image features, enabling efficient and robust modeling of temporal information. %SOLOFusion \cite{park2022time} proposed creating a cost volume from a long history of image observations for temporal multi-view 3D object detection, enhancing object perception through optimized multi-view matching configurations. CAPE \cite{cape_xiong_2023} constructs 3D position embeddings within the local camera-view coordinate system rather than the global coordinate system, allowing the 3D position embedding to be independent of camera extrinsic parameters. It addresses multi-view 3D object detection by leveraging object queries from previous frames and encoding ego motion to enhance detection. BEVDepth \cite{li2023bevdepth} prioritizes precise depth estimation to enhance the reliability and accuracy of object detection, providing a 3D object detection system specifically designed for camera-based Bird's-Eye-View (BEV) 3D object detection.  Additionally, BEVStereo \cite{liBevstereoEnhancingDepth2023} is proposed to improve depth estimation in multi-view 3D object detection using dynamic temporal stereo techniques. 
Recently, StreamPETR \cite{wang2023exploring} %is a long-sequence multi-view 3D object detection framework that 
efficiently models temporal data and object tracking through frame-by-frame query propagation and motion-aware layer normalization.

\subsection{State Space Model}

State space models have emerged as a promising alternative to traditional sequence modeling approaches. 
%such as CNN \cite{krizhevsky2012imagenet}, RNN \cite{rumelhart1986learning} and Transformer \cite{vaswani2017attention}. 
One key issue with traditional attention mechanisms~\cite{vaswani2017attention,dao2022flashattention,dao2023flashattention,peng2023rwkv}, is their quadratic time and space complexity in relation to sequence length. 
To overcome this limitation, 
%\cite{random_peng_2021} introduced Random Feature Attention (RFA), a linear time and space attention mechanism that approximates the softmax function using random feature methods. In the realm of sequence modeling, 
\cite{combining_gu_2021} proposed the Linear State-Space Layer (LSSL), a model inspired by control systems that combines recurrent neural networks, temporal convolutions, and neural differential equations. \cite{efficiently_gu_2021} introduced the Structured State Space sequence model (S4), which offers a more efficient computation method while maintaining theoretical strengths for long sequence modeling tasks. Further advancements in state space models include the introduction of the S5 layer \cite{simplified_smith_2022}, 
%which builds upon the design of the S4 layer to enhance sequence modeling performance. Additionally, 
exploring the use of Gated State Spaces for long-range language modeling \cite{long_mehta_2022}. 
%showcasing the effectiveness of state space models in capturing complex dependencies in sequence classification tasks. 
Recently, the generic language model backbone, Mamba \cite{gu2023mamba}, outperforms Transformers at various sizes on large-scale real data and enjoys linear scaling in sequence length. 
These advancements have been integrated into larger representation models \cite{ma2022mega,fu2022hungry,raw_goel_2022,selective_wang_2023,localmamba_huang_2024}, further demonstrating the versatility and scalability of structured state space models in various applications. 
%For instance, \cite{ma2022mega} introduced Mega, a gated attention mechanism equipped with moving average to incorporate position-aware local dependencies into the attention mechanism, showcasing significant improvements over existing sequence models. In the context of audio generation, \cite{raw_goel_2022} proposed SaShiMi, a multi-scale architecture for waveform modeling based on the S4 model. \cite{selective_wang_2023} introduced Selective S4 (S5) for long-form video understanding, demonstrating the importance of adaptively selecting informative tokens for efficient and accurate modeling of spatiotemporal dependencies in videos. 
State space models have also been extended in visual tasks. \cite{islam2022long} uses 1D S4 to handle the long-range temporal dependencies for video classification. TranS4mer \cite{islam2023efficient} combines the strengths of S4 and self-attention, achieving state-of-the-art performance for movie scene detection. %LocalMamba \cite{localmamba_huang_2024} uses a windowed selective scan strategy to capture local dependencies in images while maintaining a global perspective. 
And VMamba \cite{liu2024vmamba} introduces a vision backbone with linear time complexity that integrates Visual State-Space (VSS) blocks with the 2D Selective Scan (SS2D) module. 

% In conclusion, state space models have shown great potential in addressing the challenges of long sequence modeling across different domains. These models offer efficient and effective solutions for capturing complex dependencies in sequential data, showcasing advancements in performance and scalability compared to traditional sequence modeling approaches. 

\begin{figure*}
\centering
\includegraphics[width=1\linewidth]{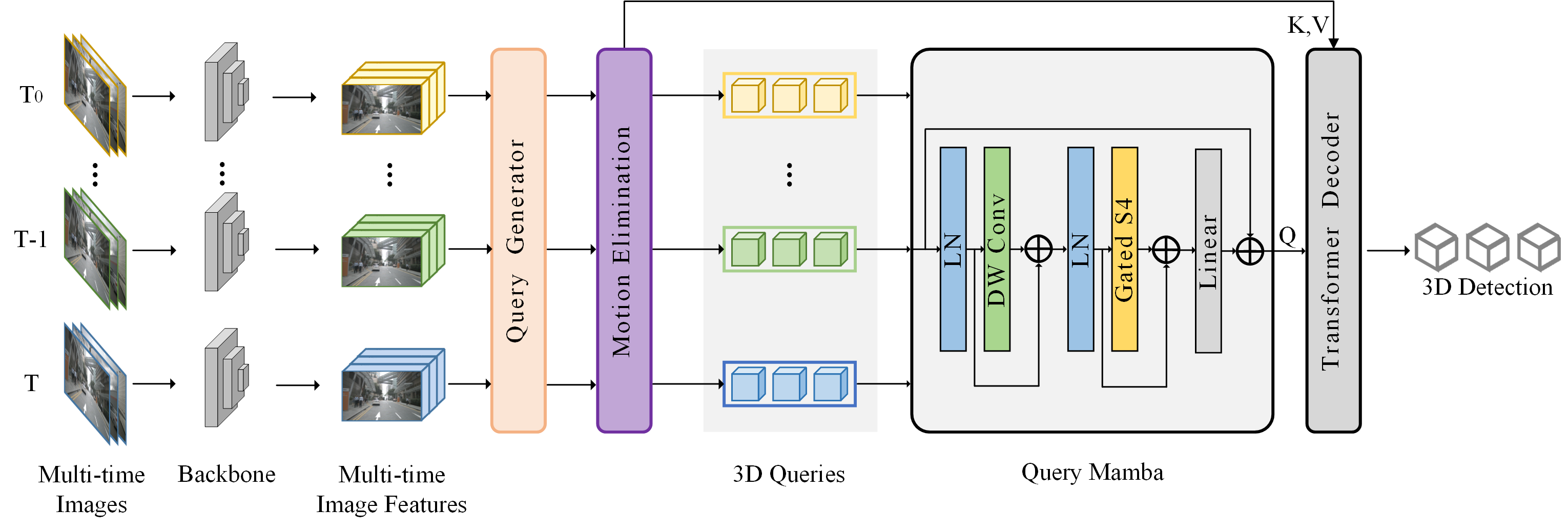} % Reduce the figure size so that it is slightly narrower than the column. Don't use precise values for figure width.This setup will avoid overfull boxes.
% \captionsetup{justification=centerlast} % Center the caption
\caption{
Overall architecture of the proposed MambaDETR. Key enhancements include 2D-priors-based query initialization, Motion Elimination to retain only moving 3D queries across frames, and Query Mamba for state-space temporal fusion. The 3D queries interact with the current image frame in a transformer decoder, producing the final 3D object detections.
% Overall architecture of DinoQuery. The image backbone network extracts features from the input multi-view images, which are then shared with the Heatmap Head and Grounding DINO modules. The Heatmap Head generates 2D global queries, while Grounding DINO produces 2D category-customized queries based on visual and textual prompts. These queries are combined to form a comprehensive set of 2D queries, which are then processed by the 3D Query Generator. This generator constructs 3D queries that are refined and passed through the decoder layers to predict the 3D detection boxes.
}
\label{pipeline}
\end{figure*}

\section{Method}
\label{sec:method}
Figure \ref{pipeline} illustrates the overall structure of the proposed MambaDETR, which follows the architecture of DETR3D \cite{wang2022detr3d}, where objects are represented as queries extracted from multi-view image features. MambaDETR develop DETR3D with the following designs: 2D-priors-based query initialization (Section \ref{Query Generator}), Motion Elimination (Section \ref{Motion Elimination}), and Query Mamba (Section \ref{Query Mamba}).
For the 2D-priors-based query initialization, we input image features and leverage a 2D detector to obtain 2D proposals. The 3D queries are then initialized from these 2D proposals through 3D projection. 
% To explore temporal information, MambaDETR extends the 2D-priors-based query initialization by retaining the 3D queries from previous frames, obtaining a sequence of 3D queries at each time step. 
The Motion Elimination Module removes redundant, motionless 3D queries from previous frames, retaining only moving 3D queries for temporal fusion.
For the Query Mamba, we leverage the 3D query sequence from multiple time steps as input and realize temporal fusion in the state-space. After that, the output queries are interact with current image frame in the transformer decoder and generate final 3D prediction.

\begin{figure*}
\centering
\includegraphics[width=1\linewidth]{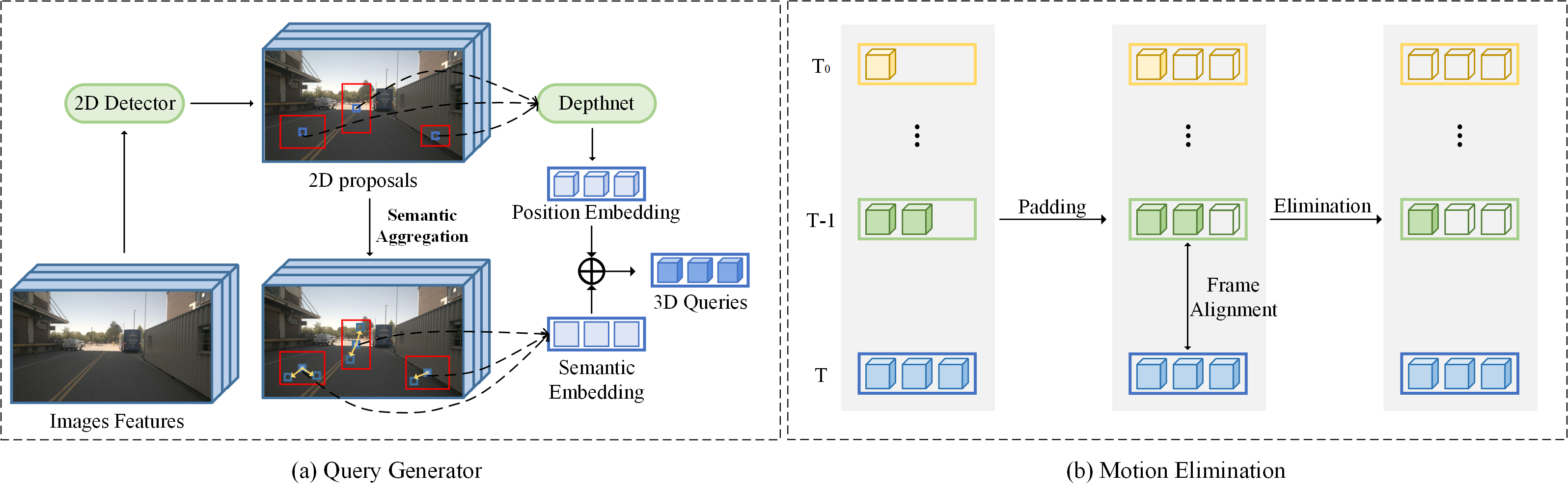} % Reduce the figure size so that it is slightly narrower than the column. Don't use precise values for figure width.This setup will avoid overfull boxes.
% \captionsetup{justification=centerlast} % Center the caption
\caption{
Detail structure of Query Generator and Motion Elimination modules in the MambaDETR architecture. (a) The Query Generator utilizes a 2D detector and DepthNet to create 2D proposals, which are transformed into 3D queries by integrating position and semantic embeddings. (b) The Motion Elimination module then filters out static 3D queries across frames by measuring the distance between the center points.
}
\label{Motion+query}
\end{figure*}

\subsection{Query Generator}
\label{Query Generator}
Previous method \cite{xie2023sparsefusion} generate numerous queries by applying maximum pooling operations on image heatmaps, which is inefficient and computationally expensive. To address this issue, existing query-based 3D object detection methods \cite{jiang2024far3d, wang2023object, ji2024enhancing} have introduced 2D proposals to enhance 3D detection performance. Inspired by these methods, the proposed 2D Priors-based Query Generator uses a 2D object detector to localize objects within specific regions. As a result, the 2D detector not only improves computational efficiency by reducing the number of queries but also provides valuable 2D priors for localizing objects in 3D space. To the best of our knowledge, MambaDETR is the first 3D object detection method that leverages a 2D detector to enhance temporal fusion performance.

Specifically, given the image features after the image backbone, we feed them into the Faster-RCNN detector \cite{wang2023object} and a lightweight depth estimation network \cite{li2023bevdepth}, resulting in a set of 2D bounding boxes and depth distributions. The 2D detector head follows the original design, while the depth distribution is represented in discretized bins \cite{jiang2024far3d}. Given each 2D bounding box in view $i$, we initialize the 2D query $\mathbf{Q}_\text{2D}$ from the center point $\mathbf{c}_{\text{2D}}=(c_\text{w}, c_\text{h})$. To aggregate semantic information of the 2D bounding box, we interact the $\mathbf{Q}_\text{2D}$ with the surrounding pixels through deformable attention, following the similar process in \cite{zhu2020deformable, xia2022vision, li2023dfa3d, liang2022recurrent}, to obtain the semantic embedding $\mathbf{Q}_\text{sem}$. The pixel candidates are chosen based on the location of the query and the sampling offsets from the image feature $\mathbf{F}$ in view $i$. The overall process is defined as follows:
\begin{multline}
        \operatorname{DeformAttn}(\mathbf{Q}_\text{2D}, \mathbf{c}_{\text{2D}}, \mathbf{F})   \\
        = \sum_{m = 1}^{N_{\text{head}}} \mathbf{W}_m\sum_{n = 1}^{N_{\text{key}}} A_{mn} \mathbf{W}'_{m} \mathbf{F} (\mathbf{c}_{\text{2D}} + \Delta \mathbf{c}_{mn}),
\end{multline}
where $\mathbf{Q}$, $\mathbf{c}_{\text{2D}}$, and $\mathbf{F}$ represent the query, reference point, and image features, respectively. The index $m$ denotes the attention head, and $N_{\text{head}}$ 
is the total number of attention heads. The index $n$ corresponds to the sampled key, and 
$N_{\text{key}}$ is the total number of sampled keys for each head. $A_{mn} \in [0, 1]$ represents the predicted attention weight and $\mathbf{W}_m$, $\mathbf{W}'_{m}$ are the learnable weights. The vector $\Delta \mathbf{c}_{mn} \in \mathbb{R}^2$ represents the predicted offsets to the reference point $\mathbf{c}_{\text{2D}}$. The term $\mathbf{F}(\mathbf{c}_{\text{2D}} + \Delta \mathbf{c}_{mn})$ represents the feature at location $\mathbf{c}_{\text{2D}} + \Delta \mathbf{c}_{mn}$.

To further lift the 2D query into the 3D space, the center point $\mathbf{c}_{\text{2D}}=(c_\text{w}, c_\text{h})$ of each 2d bounding box is combined with the corresponding predicted depth distribution $\mathbf{d}_{\text{c}}$ and project to the 3D center points $\mathbf{c}_{\text{3D}}$ of 3D proposals:
\begin{equation}
         \mathbf{c}_{\text{3D}} = \mathbf{K}^{-1}_j \mathbf{I}^{-1}_j [\mathbf{c}_\text{w} \ast \mathbf{d}_{\text{c}}, \mathbf{c}_\text{h} \ast \mathbf{d}_{\text{c}}, \mathbf{d}_{\text{c}}, \mathbf{1}]^T,
\end{equation}
where $\mathbf{K}_j$, $\mathbf{I}_j$ denote extrinsic and intrinsic matrices of the $j$-th camera.

After generating center points $\mathbf{c}_{\text{3D}}$ of the 3D proposals, we use sinusoidal transformation and MLP to obtain the 3D position embedding $\mathbf{Q}_\text{pos}$ following the process and combine with the aforementioned $\mathbf{Q}_\text{sem}$ together and result in 3D query $\mathbf{Q}_{\text{3D}}$:
\begin{equation}
\begin{aligned}
         \mathbf{Q}_{\text{pos}} &= \operatorname{PosEmbed}(\mathbf{c}_{\text{3D}}), \\
         \mathbf{Q}_{\text{3D}} &= \mathbf{Q}_{\text{pos}} + \mathbf{Q}_{\text{sem}}.
\end{aligned}
\end{equation}

\subsection{Motion Elimination}
\label{Motion Elimination}
The Motion Elimination (ME) module is applied optimize the computation cost by discarding the relative static object queries for the subsequent temporal fusion. In more detail, we present the structure of ME module in Figure \ref{Motion+query}b, which consists of three parts: zero padding, frame alignment.
For the zero padding part, given the 3D queries sequence $\mathbf{Q}^\text{seq}_{\text{3D}} = [\mathbf{Q}^{t}_{\text{3D}}, \mathbf{Q}^{t-1}_{\text{3D}}, \cdots ,\mathbf{Q}^{t-N}_{\text{3D}}]$, 
we select the 3D queries with the in the $i-th$ frame $\mathbf{Q}^i_{\text{3D}}$ that have the largest number of 3D proposal and represented as $K$. Then we add zero padding to all the 3D queries in the remaining frames and get the updated 3D queries sequence $\mathbf{\hat{Q}}^\text{seq}_{\text{3D}} = [\mathbf{\hat{Q}}^{t}_{\text{3D}}, \mathbf{\hat{Q}}^{t-1}_{\text{3D}}, \cdots ,\mathbf{\hat{Q}}^{t-N}_{\text{3D}}]$ with size of $K \times N$.  

For the frame alignment, we select the current queries $\mathbf{\hat{Q}}^{t}_{\text{3D}}$, previous queries in the $t$-$j$  frame $\mathbf{\hat{Q}}^{t-j}_{\text{3D}}$ and their corresponding 3D center points $\mathbf{C}^{t}_{\text{3D}}$, $\mathbf{C}^{t-j}_{\text{3D}}$. Then the previous query $\mathbf{\hat{Q}}^{t-j}_{\text{3D}}$ is aligned to the current frame through ego transformation, which following the process in \cite{hou2024query}. Given the ego pose matrix of $\mathbf{R}^t_{w}$ and $\mathbf{R}^{t-j}_{w}$ in the current and $t$-$j$  frame, we first compute the ego transformation matrix and align the center of the objects in the $t$-$j$  frame to the current frame. The overall process is formulated as:
\begin{equation}
\begin{aligned}
         \mathbf{R}^t_{t-j} &= \mathbf{R}^t_{\text{w}} \cdot \operatorname{inv}(\mathbf{R}^{t-j}_{w}), \\
         \mathbf{C}'_{t-1} &= (\mathbf{C}_{t-1} + \mathbf{V}_{t-j} \cdot \Delta t) \cdot (\mathbf{R}^t_{t-j}) ^T.
\end{aligned}
\end{equation}
After getting the aligned center points, we utilize the L2 center distance of object and create cost matrix $\mathbf{D}^t_{t-j} \in \mathbb{R}^{K \times K}$ to measure the moving distance of objects between the chosen frames. 
\begin{equation}
         \mathbf{D}^t_{t-j} = \mathbf{L}_2 (\mathbf{C}_{t} - \mathbf{C}'_{t-1}) \\
\end{equation}
Then we distinguish the relative static position of the objects with two standards: The L2 distance of object is below the threshold $\alpha$ and the category of the objects in frame $t$ and $t$-$j$ are the same.
\begin{equation}
\begin{aligned}
    \mathbf{M}_{t-j} &= \left\{ 
    \begin{array}{ll}
    0, & D^t_{t-j} \leq \alpha \text{ and } \text{CAT}_t = \text{CAT}_{t-j} \\
    1e^6, & D^t_{t-j} > \alpha \text{ or } \text{CAT}_t \neq \text{CAT}_{t-j}
    \end{array}
    \right., \\
    \mathbf{A}_{t-j} &= \operatorname{softmax}(\mathbf{M}_{t-j}),
\end{aligned}
\end{equation}
where $\mathbf{A}_{t-j} \in \mathbb{R}^{K \times K}$. We identify the object of $t$-$j$ frame is relative static if it has one relative static relationship in the $t$ frame. The process is formulated as:
% $\mathbf{B}_j \in \mathbb{R}^{1 \times K}$
\begin{equation}
B^{n}_j = 
\begin{cases} 
1 & \text{if } \sum_{m=1}^{K} \mathbf{A}^{mn}_{t-j} = K \\
0 & \text{otherwise}
\end{cases}.
\end{equation}
% \begin{equation}
% % \mathbf{B}_{t-j} = [B^{0}_{t-j}, \cdots, B^{n}_{t-j}, \cdots, B^{K}_{t-j}] , \quad \text{for } n = 1, 2, \ldots, K
%  \mathbf{B}_{t-j} = \{ B^{n}_{t-j} \}^{K}_{n=1}
% \end{equation}
After that, we can obtain the motion mask of the $t$-$j$ frame $\mathbf{B}_{t-j} = \{ B^{n}_{t-j} \}^{K}_{n=1} \in \mathbb{R}^{1 \times K}$.
% where $\mathbf{B}_{t-j} \in \mathbb{R}^{1 \times K}$ is the motion mask of the $t$-$j$ frame.
We extend the above process to all the previous frames and gather the motion mask $\mathbf{B} = [\mathbf{B}^T_t, \mathbf{B}^T_{t-1}, \cdots , \mathbf{B}^T_{t-N}]$. To be notice that we set all the elements of $\mathbf{B}_t$ to be 1, which means all the queries of the current frame will be remained. Finally, we multiply the 3D queries sequence with the motion mask and get the relative moving query sequence:
\begin{equation} 
    \begin{aligned}
         \mathbf{Q}_{\text{moving}} &= \mathbf{\hat{Q}}^\text{seq}_{\text{3D}} \ast \mathbf{B} \\
         &= [\mathbf{\hat{Q}}^{t}_{\text{3D}}, \mathbf{\hat{Q}}^{t-1}_{\text{3D}}, \cdots ,\mathbf{\hat{Q}}^{t-N}_{\text{3D}}] \ast  [\mathbf{B}^T_t, \mathbf{B}^T_{t-1}, \cdots , \mathbf{B}^T_{t-N}].
    \end{aligned}
\end{equation}

\subsection{Query Mamba}
\label{Query Mamba}
\subsubsection{Background}
The Structured State-Space Model (SSM) can be regarded as linear-time-invariant system that map the input stimulation $x(t)$ to the response $y(t)$ through the hidden state $h(t) \in \mathbb{R}^{N}$. The SSM can be defined in the continuous time using the following equations:
\begin{equation} 
\begin{aligned}
         \mathbf{h}'(t) &= \mathbf{A}\mathbf{h}(t) + \mathbf{B}x(t), \\
    y(t) &= \mathbf{C}\mathbf{h}(t) + Dx(t),
\end{aligned}
\end{equation}
where $\mathbf{A} \in \mathbb{R}^{N \times N}, \mathbf{B} \in \mathbb{R}^{N \times 1}, \mathbf{C} \in \mathbb{R}^{1 \times N}$ and $D  \in \mathbb{R}^{1}$ are the weighting parameters.

To apply the continuous SSM into the deep models, we introduce a timescale parameter $\Delta$ and discretize the SSM as follow:
\begin{equation}  
\begin{aligned}
    \mathbf{h}_k &= \bar{\mathbf{A}} \mathbf{h}_{k-1} + \bar{\mathbf{B}} \mathbf{x}_k, \\
    \mathbf{y}_k &= \bar{\mathbf{C}} \mathbf{h}_k + \bar{D} \mathbf{x}_k,
    \label{eq:DSSM}
\end{aligned}
\end{equation}
where $\bar{\mathbf{A}}, \bar{\mathbf{B}}, \bar{\mathbf{C}}$ and $\bar{D}$ are the discrete version of $\mathbf{A} , \mathbf{B} , \mathbf{C}$ and $D$. The equation \eqref{eq:DSSM} can also be expressed as a convolution operation:
\begin{equation}
\begin{aligned}
\bar{\mathbf{K}} &= \left( \bar{\mathbf{C}}\bar{\mathbf{B}}, \bar{\mathbf{C}}\bar{\mathbf{A}}\bar{\mathbf{B}}, \ldots, \bar{\mathbf{C}}\bar{\mathbf{A}}^{L-1}\bar{\mathbf{B}} \right), \\
    \mathbf{y} &= \bar{\mathbf{K}} * \mathbf{x},
    \end{aligned}
\end{equation}
where $\bar{\mathbf{K}}$ is convolution kernel and $L$ is the length of the input sequence.

\subsubsection{Query Mamba}
The overview of the proposed Query Mamba is shown in Figure \ref{pipeline}. For the queries $\mathbf{\hat{Q}}^{t-i}_{\text{3D}}$ in the $t$-$i$ frame, we concatenate all the queries along the channel dimension $D$ and get the fusion candidate $\mathbf{Q}^{t-i}_{\text{moving}} \in \mathbb{R}^{1 \times KD}$. After that, following the standard 1-D input of Mamba, we send the fusion candidate sequence $\mathbf{Q}_{\text{moving}} = \{ \mathbf{Q}^{t-i}_{\text{moving}} \}^{N}_{i=1} \in \mathbb{R}^{N \times KD}$ into the Query Mamba module.

First, we pass the sequence through the Layer Normalization layer. Subsequently, we apply a depth-wise convolution (DW Conv) along with a residual connection, which improves the efficiency of the CNN and has been used in VMamba \cite{liu2024vmamba}. Afterward, we apply another Layer Normalization and a Gated S4 (GS4) layer, also with a residual connection. Finally, the sequence is passed through a Linear layer with a residual connection that connects from the initial input of Query Mamba:
\begin{align*}
\mathbf{z} &= \operatorname{DWConv}(\operatorname{LN}(\mathbf{Q}_{\text{moving}}))+\operatorname{LN}(\mathbf{Q}_{\text{moving}}), \\
\mathbf{z}' &= \operatorname{GS4}(\operatorname{LN}(\mathbf{z}))+\operatorname{LN}(\mathbf{z}), \\
\mathbf{Q}_{\text{out}} &= \operatorname{Linear}(\mathbf{z}') + \mathbf{Q}_{\text{moving}}.
\end{align*}

The SSM-based model plays a vital role in our Mamba DETR method, so as the cross-attention in the transformer-based temporal fusion methods \cite{wang2023exploring, li2022bevformer}. Given the input sequence $\mathbf{Q}_{\text{moving}} = \{ \mathbf{Q}^{t-i}_{\text{moving}} \}^{N}_{i=1} \in \mathbb{R}^{N \times KD}$, the computational complexity of cross-attention and SSM is formulated as:
\begin{align*}
\Omega(\text{cross-attention}) &= 4N(KD)^2 + 2N^2KD, \\
\Omega(\text{SSM}) &= 3N(2D)M + N(2KD)M,
\end{align*}
where $M$ is a fixed parameter set to 16. As we can see, the computational complexity of cross-attention is quadratic in the sequence length $N$, while the SSM is linear in $N$. This computational advantage makes the SSM-based method scalable for long sequence modeling and thus beneficial for long-range exploration.
\section{Experiments}
\label{sec:exp}

\begin{table}[ht]
\caption{Ablation study on different components of MambaDETR}\label{ablation}
	\centering
	\scalebox{1}{
\begin{tabular}{ c c c c| c c }
%\toprule
 \hline
 2D detector & SE & ME  & QMamba & NDS & mAP\\
 \hline
 \tikzxmark & \tikzxmark & \tikzxmark & \tikzxmark & 57.5 & 49.4 \\  
\tikzcmark  & \tikzxmark & \tikzxmark  & \tikzxmark & 58.1 & 49.8 \\ 
\tikzcmark & \tikzcmark  & \tikzxmark  & \tikzxmark & 58.3 & 50.4 \\
 \tikzcmark & \tikzcmark & \tikzcmark & \tikzxmark & 58.6 & 50.5  \\
 \tikzcmark & \tikzcmark & \tikzcmark & \tikzcmark & \textbf{60.5} & \textbf{50.8}  \\
 \hline
 %\bottomrule
\end{tabular}}
\end{table}

\begin{table}[ht]
\caption{Ablation Study on the Computational Efficiency of the 2D Detector and Motion Elimination}\label{ablation_query}
	\centering
	\scalebox{1}{
\begin{tabular}{ c c | c c }
%\toprule
 \hline
 2D Detector & ME   & FPS & Memory (GB)\\
 \hline
 \tikzxmark & \tikzxmark  & 30.4 & 16.0 \\  
\tikzcmark  & \tikzxmark   & 29.7 & 16.1 \\ 
\tikzcmark & \tikzcmark   & \textbf{31.9} & \textbf{15.5} \\
 \hline
 %\bottomrule
\end{tabular}}
\end{table}

\begin{table}[h]
    \centering
    \begin{minipage}[t]{0.2\textwidth}
        \centering
        \captionof{table}{S4 variants}\label{S4_type}
        \begin{tabular}{cc}
            \hline
            S4 variant & mAP \\
            \hline
            S4 & 50.1 \\
            S6 & 50.3 \\
            DS4 & \textbf{50.8} \\
            GS6 & 49.4 \\
            \hline
        \end{tabular}
    \end{minipage}%
    \begin{minipage}[t]{0.2\textwidth}
        \centering
        \captionof{table}{Number of layers}\label{S4_number}
        \begin{tabular}{cc}
            \hline
            S4 layers & mAP \\
            \hline
            1 & 49.1 \\
            2 & 49.7 \\
            4 & 50.5 \\
            6 & \textbf{50.8} \\
            \hline
        \end{tabular}
    \end{minipage}
\end{table}

\begin{table*}
\caption{Comparison with existing methods on nuScenes \emph{validation} set.}\label{benchmark_val}
	\centering
	\setlength{\tabcolsep}{3pt}
	\begin{tabular}{c|c|c|c c|c|c|c|c|c}
	 \hline
	 & \textbf{Backbone}  & \textbf{Frames} & \textbf{NDS} $\uparrow$ & \textbf{mAP} $\uparrow$ & \textbf{mATE} $\downarrow$ & \textbf{mASE} $\downarrow$ & \textbf{mAOE} $\downarrow$  & \textbf{mAVE} $\downarrow$ & \textbf{mAAE} $\downarrow$ \\
	 \hline
	 BEVDet \cite{huang2021bevdet}           & ResNet50  & 1      & 37.9 & 29.8 & 72.5 & 27.9 & 58.9 & 86.0 & 24.5 \\
	 BEVDet4D \cite{huang2022bevdet4d}       & ResNet50  & 2      & 45.7 & 32.2 & 70.3 & 27.8 & 49.5 & 35.4 & 20.6 \\
	 PETRv2 \cite{liu2023petrv2}           & ResNet50  & 2      & 45.6 & 34.9 & 70.0 & 27.5 & 58.0 & 43.7 & 18.7 \\
	 BEVDepth \cite{li2023bevdepth}        & ResNet50  & 2      & 47.5 & 35.1 & 63.9 & 26.7 & 47.9 & 42.8 & 19.8 \\
	 BEVStereo \cite{liBevstereoEnhancingDepth2023}       & ResNet50  & 2      & 50.0 & 37.2 & 59.8 & 27.0 & 43.8 & 36.7 & 19.0 \\
	 BEVFormer v2 \cite{yangBevformerV2Adapting2023b}     & ResNet50  & -      & 52.9 & 42.3 & 61.8 & 27.3 & 41.3 & 33.3 & 18.8 \\
	 SOLOFusion \cite{park2022time}      & ResNet50  & 16+1   & 53.4 & 42.7 & 56.7 & 27.4 & 51.1 & 25.2 & 18.1 \\
	 BEVPoolv2 \cite{huangBevpoolv2CuttingedgeImplementation2022b}        & ResNet50  & 8+1    & 52.6 & 40.6 & 57.2 & 27.5 & 46.3 & 27.5 & 18.8 \\

	 StreamPETR \cite{wang2023exploring}          & ResNet50  & 8      & 55.0 & 45.0 & 61.3 & 26.7 & 41.3 & 26.5 & 19.6 \\
	 DETR3D \cite{wang2022detr3d}         & ResNet101-DCN & 1      & 43.4 & 34.9 & 71.6 & 26.8 & 37.9 & 84.2 & 20.0 \\
	 Focal-PETR \cite{wang2023focal}      & ResNet101-DCN & 1      & 46.1 & 39.0 & 67.8 & 26.3 & 39.5 & 80.4 & 20.2 \\
	 PETR \cite{petr_liu_2022}           & ResNet101-DCN & 1      & 44.1 & 36.6 & 71.7 & 26.7 & 41.2 & 83.4 & 19.0 \\
	 BEVFormer \cite{li2022bevformer}       & ResNet101-DCN & 4      & 51.7 & 41.6 & 67.3 & 27.4 & 37.2 & 39.4 & 19.8 \\
	 PolarDETR \cite{chenPolarParametrizationVisionbased2022c}      & ResNet101-DCN & 2      & 48.8 & 38.3 & 70.7 & 26.9 & 34.4 & 51.8 & 19.6 \\
	 Sparse4D \cite{lin2022sparse4d}        & ResNet101-DCN & 4      & 54.1 & 43.6 & 63.3 & 27.9 & 36.3 & 31.7 & 17.7 \\
	 BEVDepth \cite{li2023bevdepth}             & ResNet101 & 2      & 53.5 & 41.2 & 56.5 & 26.6 & 35.8 & 33.1 & 19.0 \\
	 SOLOFusion \cite{park2022time}           & ResNet101 & 16+1   & 58.2 & 48.3 & \textbf{50.3} & 26.4 & 38.1 & 24.6 & 20.7 \\
	 StreamPETR \cite{wang2023exploring}          & ResNet101 & 8      & 59.2 & 50.4 & 56.9 & \textbf{26.2} & 31.5 & 25.7 & 19.9 \\
  MambaDETR          & ResNet101 & 8      & \textbf{60.5} & \textbf{50.8} & 53.7 & 26.4 & \textbf{30.7} & \textbf{24.4} & \textbf{17.2} \\
	 \hline
	\end{tabular}
\end{table*}

\begin{table*}
\caption{Comparison with existing methods on nuScenes \emph{test} set.}\label{benchmark_test}
    \centering
    \begin{tabular}{c|c|c|c|c|c|c|c|c}
     \hline
     & \textbf{Backbone}  & \textbf{NDS} $\uparrow$ & \textbf{mAP} $\uparrow$ & \textbf{mATE} $\downarrow$ & \textbf{mASE} $\downarrow$ & \textbf{mAOE} $\downarrow$  & \textbf{mAVE} $\downarrow$ & \textbf{mAAE} $\downarrow$ \\
     \hline
     FCOS3D \cite{wangFcos3dFullyConvolutional2021b}          & R101-DCN     & 35.8 & 42.8 & 69.0 & 24.9 & 45.2 & 143.4 & 12.4 \\
     DETR3D \cite{wang2022detr3d}          & V2-99        & 41.2 & 47.9 & 64.1 & 25.5 & 39.4 & 84.5 & 13.3 \\
     MV2D \cite{wang2023object}            & V2-99        & 46.3 & 51.4 & 54.2 & 24.7 & 40.3 & 85.7 & 12.7 \\
     UVTR \cite{liUnifyingVoxelbasedRepresentation2022c}            & V2-99        & 47.2 & 55.1 & 57.7 & 25.3 & 39.1 & 50.8 & 12.3 \\
     BEVFormer \cite{li2022bevformer}       & V2-99        & 48.1 & 56.9 & 58.2 & 25.6 & 37.5 & 37.8 & 12.6 \\
     PETRv2 \cite{liu2023petrv2}          & V2-99        & 49.0 & 58.2 & 56.1 & 24.3 & 36.1 & 34.3 & 12.0 \\
     PolarFormer \cite{jiang2023polarformer}     & V2-99        & 49.3 & 57.2 & 55.6 & 25.6 & 30.4 & 43.9 & 12.7 \\
     BEVStereo \cite{liBevstereoEnhancingDepth2023}       & V2-99        & 52.5 & 61.0 & \textbf{43.1} & 24.6 & 35.8 & 35.7 & 13.8 \\
     StreamPETR \cite{wang2023exploring}          & V2-99        & 55.0 & 63.6 & 47.9 & 23.9 & 31.7 & 24.1 & 11.9 \\
     BEVDet4D \cite{huang2022bevdet4d}        & Swin-B [32]  & 45.1 & 56.9 & 51.1 & 24.1 & 38.6 & 30.1 & \textbf{12.1} \\
     BEVDepth \cite{li2023bevdepth}        & ConvNeXt-B   & 52.0 & 60.9 & 44.5 & 24.3 & 35.2 & 34.7 & 12.7 \\
     AeDet \cite{aedet_feng_2022}           & ConvNeXt-B   & 53.1 & 62.0 & 43.9 & 24.7 & 34.4 & 29.2 & 13.0 \\
     PETRv2 \cite{liu2023petrv2}              & RevCol-L [3] & 51.2 & 59.2 & 54.7 & 24.2 & 36.0 & 36.7 & 12.6 \\
     SOLOFusion \cite{park2022time}      & ConvNeXt-B   & 54.0 & 61.9 & 45.3 & 25.7 & 37.6 & 27.6 & 14.8 \\
     StreamPETR \cite{wang2023exploring}          & ViT-L [9]    & \textbf{62.0} & 67.6 & 47.0 & 24.1 & \textbf{25.8} & 23.6 & 13.4 \\
     MambaDETR           & ViT-L [9]    & 60.7 & \textbf{68.2} & 48.1 & \textbf{23.9} & 26.1 & \textbf{22.8} & 12.7 \\
     \hline
    \end{tabular}
\end{table*}

\subsection{Dataset and Metrics}
The nuScenes dataset is a comprehensive resource for 3D object detection, encompassing 1,000 scenes, each lasting about 20 seconds and annotated at a frequency of 2 Hz. This dataset features images captured from six cameras, alongside data from five radars and one LiDAR system, providing a complete 360° field of view. The annotations include up to 1.4 million 3D bounding boxes across ten categories: car, truck, bus, trailer, construction vehicle, pedestrian, motorcycle, bicycle, barrier, and traffic cone. The scenes are organized into training (700), validation (150), and testing (150) sets. For evaluation, we utilize several metrics, including the mean Average Precision (mAP), nuScenes Detection Score (NDS), and various True Positive metrics such as Average Translation Error (ATE) and Average Velocity Error (AVE). The mAP is determined based on the distance between 2D centers on the ground plane, and NDS provides a holistic measure of detection performance by aggregating other relevant indicators.

\begin{figure*}
\centering
\includegraphics[width=1\linewidth]{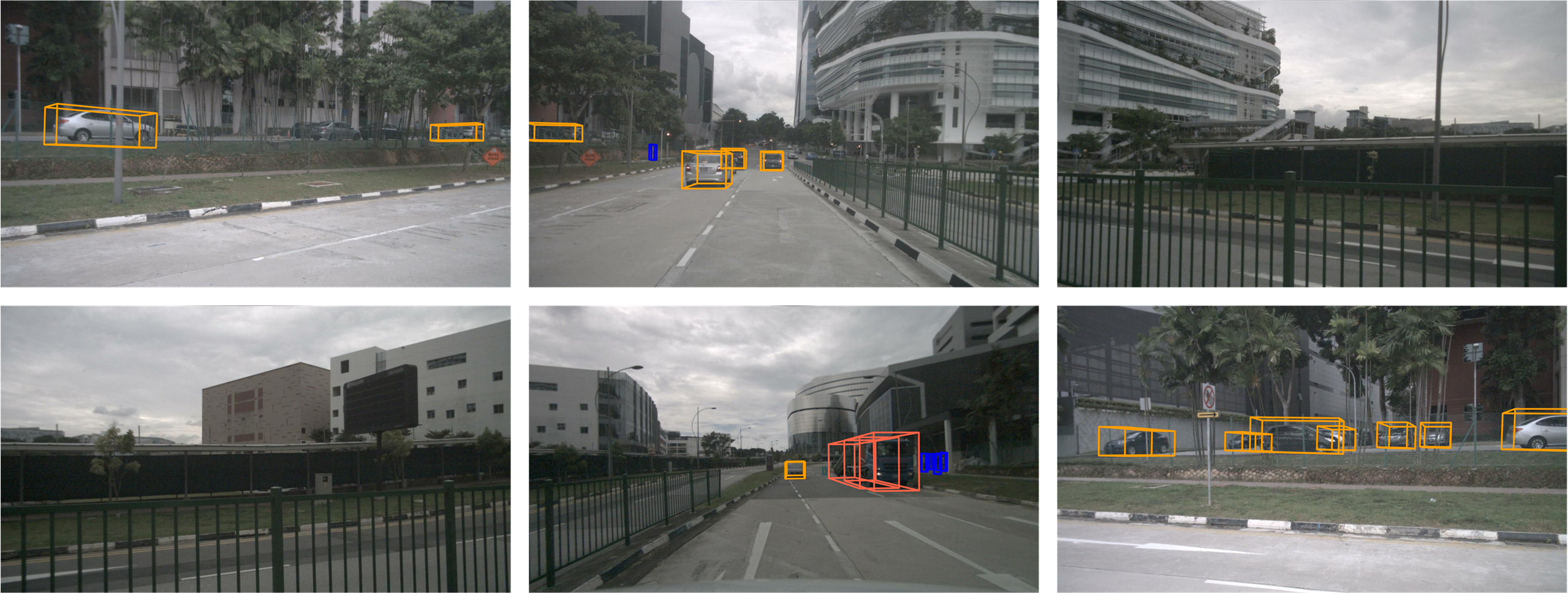} % Reduce the figure size so that it is slightly narrower than the column. Don't use precise values for figure width.This setup will avoid overfull boxes.
% \captionsetup{justification=centerlast} % Center the caption
\caption{Visualization of MambaDETR.
% Overall architecture of DinoQuery. The image backbone network extracts features from the input multi-view images, which are then shared with the Heatmap Head and Grounding DINO modules. The Heatmap Head generates 2D global queries, while Grounding DINO produces 2D category-customized queries based on visual and textual prompts. These queries are combined to form a comprehensive set of 2D queries, which are then processed by the 3D Query Generator. This generator constructs 3D queries that are refined and passed through the decoder layers to predict the 3D detection boxes.
}
\label{visualization}
\end{figure*}

\subsection{Ablation Study}
\subsubsection{2D Detector and Semantic aggregation}
We analyze the impact of the 2D detector and the Semantic Aggregation (SE) module on generating object queries, as shown in Table \ref{ablation} and Table \ref{ablation_query}. Specifically, adding the 2D detector for query generation achieves significant improvements, with mAP increasing by 0.6\% and NDS increasing by 0.4\%. This result indicates that the 2D detector can provide reliable 2D priors for the query generator and thus benefit 3D object detection.
Besides, it can also be seen that the 2D detector negatively impacts inference speed and GPU memory usage. We argue that the additional computational cost introduced by the 2D detector outweighs the savings from the reduced number of queries. Semantic aggregation is also proven to be effective in improving the quality of queries, indicating the importance of semantic information in query generation.

\subsubsection{Motion Elimination}
MambaDETR reduces the non-trivial queries through the Motion Elimination (ME) module. Table \ref{ablation_query} shows that the ME module can significantly improve inference speed by 1.2 FPS and reduce GPU memory usage by 0.6 GB, which indicates that the ME module successfully removes the non-vital queries. Furthermore, the proposed ME module also achieves a limited improvement of 0.1\% in mAP and 0.3\% in NDS. This indicates that reducing redundant queries can not only improve computational efficiency but also benefit detection performance.

\subsubsection{Query Mamba}
We experiment with different S4 variants (Table \ref{S4_type}) and the impact of varying the number of layers (Table \ref{S4_number}). In Table \ref{S4_type}, we evaluate several S4 variants, including the standard S4, S6, Diagonal S4 (DS4), and Gated S4 (GS4). Our findings indicate that GS6 achieves the highest mAP of 50.8, outperforming the other variants. In Table \ref{S4_number}, we assess the effect of increasing the number of S4 layers on mAP. The results demonstrate that performance improves steadily with additional layers, stabilizing at an mAP of 50.8 with 6 layers. The result indicates that GS4 model with 6 layer can achieves the best performance.

\subsection{Results and Analysis}
\subsubsection{Main Results}
We compare the proposed Mamba DETR with previous state-of-the-art vision-based 3D detectors on the nuScenes validation and test sets. As shown in Table \ref{benchmark_val}, Mamba DETR demonstrates superior performance on the validation set, achieving significant improvements in NDS, mAP, and localization metrics compared to other methods. Specifically, with a ResNet101 backbone and 8-frame input, Mamba DETR outperforms StreamPETR by 0.7\% in NDS and 0.4\% in mAP. The mATE value of Mamba DETR also shows notable improvements, reducing the localization error compared to most competitors. This indicates the efficacy of Mamba DETR's object detection and tracking capabilities.

When comparing the performance on the test set in Table \ref{benchmark_test}, Mamba DETR again achieves impressive results. With a ViT-Large backbone, Mamba DETR reaches an mAP of 68.2\% and an NDS of 60.7\%, surpassing StreamPETR by 0.6\% in mAP and demonstrating comparable performance in mASE and mAOE metrics. Notably, Mamba DETR also achieves lower mAVE compared to many competing methods, suggesting improved velocity estimation and better temporal consistency for tracking moving objects. 
Besides, we show the visualization of the detection results In Figure \ref{visualization}, which also effectively demonstrate the superior performance of the proposed MambaDETR.

% Overall, Mamba DETR showcases strong performance across both nuScenes validation and test sets, particularly excelling in detection accuracy and localization. The significant gains in mAP and NDS demonstrate the strength of our proposed approach, especially when utilizing advanced backbones like ResNet101 and ViT-Large. The improvements in metrics such as mAOE and mAVE further highlight the advantages of Mamba DETR in complex, dynamic environments.

\subsubsection{Temporal Extent Analysis}
To validate long-term exploration of input image sequence, we evaluate both StreamPETR and MambaDETR with different numbers of training frames (i.e., 1, 2, 4, 8, and 12). The results are presented in Figure \ref{line-chart} (a). First, we notice that a short temporal window of 2 training frames produces a suboptimal performance of 40.7\% mAP. Expanding the temporal range to 8 frames boosts the performance to 50.8\% mAP, reflecting a 10.1\% improvement. Extending further to 12 frames continues to enhance the mAP, reaching 52.2\%, which indicates that adding more frames does not obtain promising improvements. Compared to our baseline, StreamPETR, MambaDETR consistently outperforms in terms of mAP across all training frames, particularly excelling with an extended temporal window. This demonstrates the superior capability of our proposed method to leverage long-range temporal information, resulting in significantly better detection performance.

\subsubsection{Computational Efficiency Analysis}
\begin{figure}
\centering
\includegraphics[width=1\linewidth]{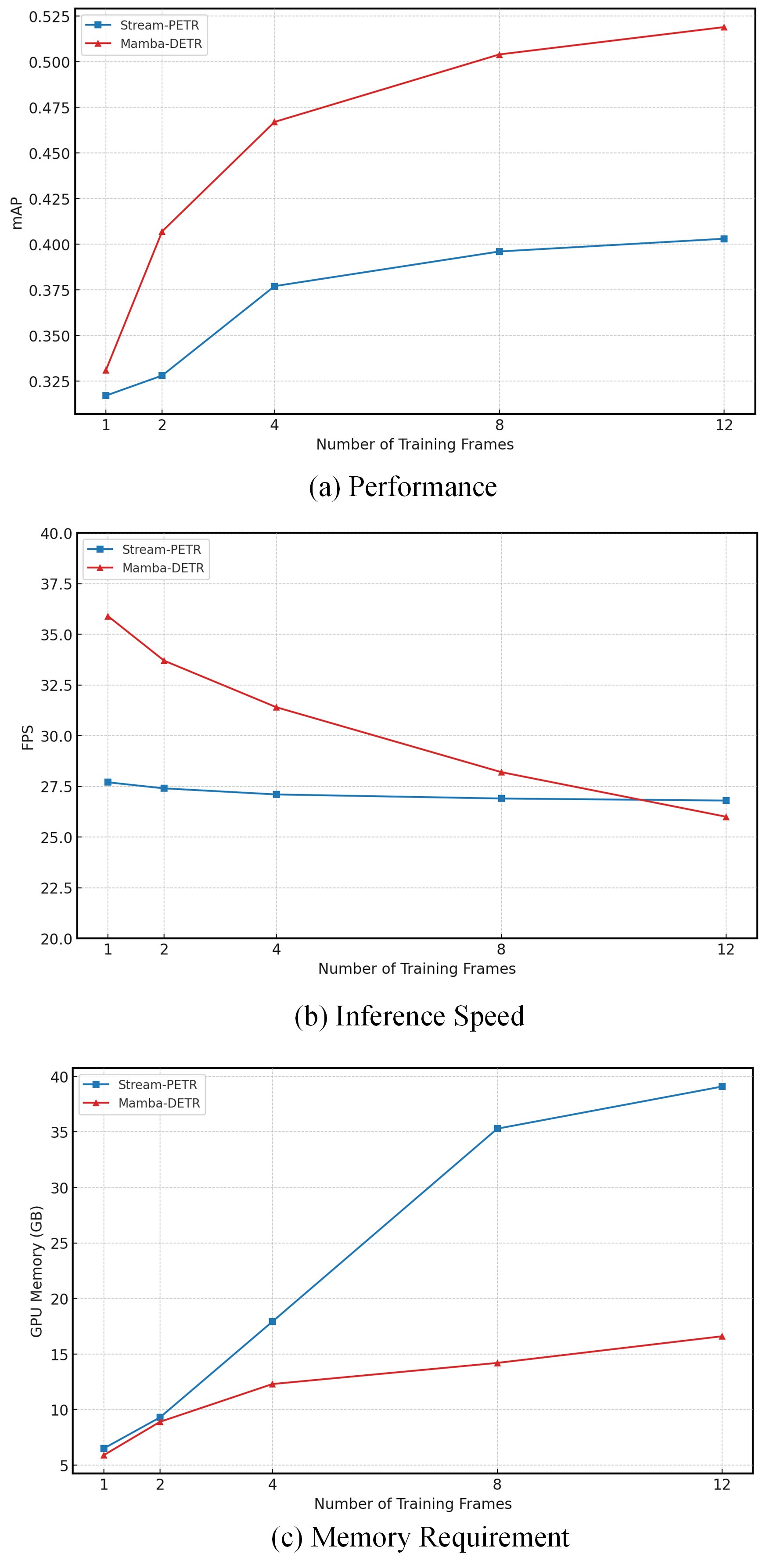} % Reduce the figure size so that it is slightly narrower than the column. Don't use precise values for figure width.This setup will avoid overfull boxes.
% \captionsetup{justification=centerlast} % Center the caption
\caption{ Performance(mAP), Inference Speed(samples/Second) and Memory Requirements(GB) of MambaDETR as the function of the sequence length.
% Overall architecture of DinoQuery. The image backbone network extracts features from the input multi-view images, which are then shared with the Heatmap Head and Grounding DINO modules. The Heatmap Head generates 2D global queries, while Grounding DINO produces 2D category-customized queries based on visual and textual prompts. These queries are combined to form a comprehensive set of 2D queries, which are then processed by the 3D Query Generator. This generator constructs 3D queries that are refined and passed through the decoder layers to predict the 3D detection boxes.
}
\label{line-chart}
\end{figure}

In Figure \ref{line-chart}b and c, we analyze the computational efficiency of the MambaDETR in two aspect: inference speed and memory cost. In Figure \ref{line-chart}b, we observe that the inference speed decreases as the sequence length increase from 35.9\% to 26.0\% in FPS. Meanwhile, the inference speed of StreamPETR remains stable around 27\% due to the recurrent temporal fusion mechanism that keep the fusion candidates invariant to the sequence length. However, our MambaDETR still infer faster than StreamPETR except for the frame numbers equal to 12. Moreover, Figure \ref{line-chart}c demonstrates that MambaDETR has significantly lower memory consumption compared to StreamPETR, with only linear memory increase as the sequence length grow. With 12 frames, MambaDETR requires around 15 GB of memory, whereas StreamPETR’s memory requirement is considerably higher, peaking at nearly 40 GB. 

\section{Conclusion}
In this paper, we propose MambaDETR, an effective 3D object detection framework for long-sequence temporal fusion. Unlike previous approaches, our method uses a state-space model (SSM) to perform efficient, sequential temporal fusion with linear memory and computational complexity. Additionally, we introduce the Motion Elimination Module, which selectively retains only moving objects, enhancing fusion efficiency. MambaDETR achieves state-of-the-art performance on the nuScenes dataset, demonstrating significant improvements in computational efficiency over transformer-based methods while maintaining accuracy. We hope MambaDETR offers valuable insights for advancing long-sequence temporal modeling in autonomous driving applications.

{
    \small
    \bibliographystyle{ieeenat_fullname}
    \bibliography{main}
}


\end{document}